%% file: frontiers_main.tex
\documentclass[utf8]{FrontiersinHarvard} 

\usepackage{url,hyperref,lineno,microtype,subcaption}
\usepackage{soul,amsmath,multirow,textcomp,bbm,xcolor,glossaries,makecell, dsfont}
\usepackage[onehalfspacing]{setspace}
\usepackage[]{acronym}
\setacronymstyle{long-short}
\def\keyFont{\fontsize{8}{11}\helveticabold }
\def\firstAuthorLast{Peque\~no-Zurro {et~al.}} 
\def\Authors{Alejandro Peque\~no-Zurro\,$^{1,2,3}$, Lyes Khacef\,$^{2,3}$, Stefano Panzeri\,$^{4}$, and Elisabetta Chicca\,$^{2,3}$}
\def\Address{$^{1}$Laboratory of Neural Computation, Istituto Italiano di Tecnologia, Genova, Italy \\
$^{2}$Bio-Inspired Circuits and Systems (BICS) Lab, Zernike Institute for Advanced Materials, University of Groningen,
9747 AG Groningen, The Netherlands.\\
$^{3}$Groningen Cognitive Systems and Materials Center (CogniGron), University of Groningen, 9747 AG Groningen, The Netherlands.\\
$^{4}$Institute for Neural Information Processing, Center for Molecular Neurobiology (ZMNH), University Medical Center Hamburg-Eppendorf (UKE), Falkenried 94, 20251 Hamburg, Germany}

\begin{document}

\onecolumn
\firstpage{1}

\title{Towards efficient keyword spotting using spike-based time difference encoders} 

\author[\firstAuthorLast ]{\Authors} 
\address{} 
\correspondance{} 

\extraAuth{}

\maketitle
\input{acro_list}


\begin{abstract}
Keyword spotting in edge devices is becoming increasingly important as voice-activated assistants are widely used. However, its deployment is often limited by the extreme low-power constraints of the target embedded systems. Here, we explore the Temporal Difference Encoder (TDE) performance in keyword spotting. This recent neuron model encodes the time difference in instantaneous frequency and spike count to perform efficient keyword spotting with neuromorphic processors. We use the TIdigits dataset of spoken digits with a formant decomposition and rate-based encoding into spikes. We compare three Spiking Neural Networks (SNNs) architectures to learn and classify spatio-temporal signals. The proposed SNN architectures are made of three layers with variation in its hidden layer composed of either (1) feedforward TDE, (2) feedforward Current-Based Leaky Integrate-and-Fire (CuBa-LIF), or (3) recurrent CuBa-LIF neurons. We first show that the spike trains of the frequency-converted spoken digits have a large amount of information in the temporal domain, reinforcing the importance of better exploiting temporal encoding for such a task. We then train the three SNNs with the same number of synaptic weights to quantify and compare their performance based on the accuracy and synaptic operations. The resulting accuracy of the feedforward TDE network (89\%) is higher than the feedforward CuBa-LIF network (71\%) and close to the recurrent CuBa-LIF network (91\%). However, the feedforward TDE-based network performs 92\% fewer synaptic operations than the recurrent CuBa-LIF network with the same amount of synapses. In addition, the results of the TDE network are highly interpretable and correlated with the frequency and timescale features of the spoken keywords in the dataset. Our findings suggest that the TDE is a promising neuron model for scalable event-driven processing of spatio-temporal patterns.

\tiny
\keyFont{ \section{Keywords:} neuromorphic computing, spiking neural networks, time difference encoder, spatio-temporal pattern recognition, energy efficiency, keyword spotting.} 
\end{abstract}

\newpage
\section{Introduction}
Keyword spotting is a representative task for the increasing demand for real-time, always-on computation driven by the ongoing IoT revolution. Edge devices such as smart home solutions, smartphones, and other wearable systems are becoming ubiquitous in our daily lives. This trend pushes intelligence into tiny embedded systems that are highly constrained in energy consumption and need to interact in real-time with their environment~\citep{Murshed_etal21, Sulieman_etal22}. This combination of low-power and low-latency requirements limits the deployment of keyword spotting systems on conventional hardware, such as CPU and GPU architectures. More generally, AI progress with current algorithms and hardware limitations is predicted to become technically unsustainable~\citep{Thompson_etal20, Thompson_etal21}, in particular, when targeting constrained embedded systems~\citep{Rabaey_etal19}. Neuromorphic hardware and systems are an emerging alternative to traditional silicon solutions due to their ability to imitate the efficiency and computational power of the brain.

\gls{SNN} models implemented in neuromorphic hardware constitute an efficient alternative to building AI systems~\citep{Khacef_etal18, Roy2019, Ivanov2022}. \glspl{SNN} are considered the third generation of neural networks~\citep{Maass_97} characterized by two main properties: (1) asynchronous computing, which exploits the sparsity of event-driven sensors, and (2) computational compartments, which may have inherent temporal dynamics in the form of leakage. Hence, compared to conventional \glspl{ANN}, \glspl{SNN} uses the temporal information of the data more efficiently~\citep{Yarga2022}. \glspl{SNN} have gotten traction in the last five years thanks to the emergence of neuromorphic hardware platforms such as IBM TrueNorth~\citep{Merolla2014} and Intel Loihi~\citep{Davies2018}, effectively demonstrating the low-latency and low-power potential in several applications such as multimodal visual-EMG hand gesture recognition~\citep{Ceolini_etal20}, tactile sensing~\citep{Muller-Cleve2022}, and keyword spotting~\citep{Blouw_etal18}.

In audio classification systems, studies have historically used \gls{HMM} with a preprocessing step based on \glspl{MFCC} for speech recognition systems~\citep{Rabiner1989, Wong2001, Pan2018}. In addition to this signal processing pipeline, several studies have shown that the use of the formant decomposition of the frequency spectrum, as used in this work, improves the performance of speech recognition algorithms~\citep{Welling1998, Prica2010, Stanek2013, Coath2014, Laszko2016}. We propose using a \gls{SNN}, which consists of a novel neural model \gls{TDE}, as a more efficient and scalable system implementation to classify the spatio-temporal patterns typical of spoken words. This development aligns with previous work that shows the potential energy efficiency capability of neuromorphic on-chip computations for these keyword spotting applications~\citep{blouw2019bench}. 

Drawing inspiration from neural codes in the brain, several works have emphasized the importance of spike timing in neurons located in sensory cortices and its relevance in behavioral discrimination tasks ~\citep{engineer2008, yang2008, Zuo2015, chong2020}. Common mechanisms for encoding sensory signals involve building the pattern of spikes in time, the code's latency from stimulus presentation, the inter-spike interval between spikes, or the phase of firing neurons~\citep{Panzeri2010,  Panzeri2014}. These time-related encoding mechanisms have been found to be present across the main sensory modalities with visual~\citep{victor2000, butts2007}, auditory~\citep{moiseff1981, schnupp2006, Kayser2010} and somatosensory~\citep{panzeri2001, jones2004, montemurro2007}. Considering the auditory modality, research has shown that millisecond-precise neural codes are crucial for processing auditory signals from the peripheral nervous system to the cortex~\citep{schnupp2006, Kayser2010}. In parallel, advances in Spiking Neural Networks (SNNs) have considered multiple time-based encoding schemes as the primary encoding mechanism of artificial sensory signals for event-based computation~\citep{auge2021}. 
The proposed \gls{TDE} model~\citep{Milde2015, Angelo2020} is a type of \gls{LIF} neuron with an extra synaptic compartment and fixed input directional connectivity. The spiking rate of the cell's output is inversely proportional to the time difference of the input's spike trains. Therefore, the firing rate at its output increases with highly correlated spike trains at its input~\citep{salinas2001, valente2021}. The neuron dynamics and the cell's spatial selectivity make it especially suitable to exploit the spatio-temporal structure of the input data. Therefore, it makes it suitable for processing spatio-temporal features from all sensory modalities such as vision~\citep{Milde2015, Angelo2020, Schoepe2021}, audition~\citep{ Coath2014, Gutierrez-Galan2022}, touch~\citep{mastella2021hardware}, and olfaction. Related cortical circuits that inspire this type of implementation can be found in animals that range from insects to mammals, e.g. visual optic flow circuits~\citep{hassenstein1956system, Barlow1965, Maisak2013} and the auditory thalamocortical processing~\citep{coath2011emergent, Sandin2020}.

The structure of this paper is as follows. We apply the \gls{TDE} neuron model in a \gls{SNN} for a speech recognition task. The initial analysis of the dataset shows a significant amount of information in the time pattern of the encoded spikes. We train \glspl{SNN} using supervised learning methods and compare the performance of three different network architectures that compare the use of the \gls{TDE} neuron model against typical networks with \gls{CuBa-LIF} neurons,  finding a similar performance of the \gls{TDE} neuron model and recurrent network of \gls{CuBa-LIF} cells. Our findings show that the network with the \gls{TDE} neuron model is more efficient, with a 92\% reduction in synaptic operations, presents a more sparse signal representation, and has better interpretability of the results.


\section{Materials and methods}
The pipeline of the proposed \gls{SNN} architecture is depicted in Figure \ref{fig:method_system}. Spoken digits were preprocessed through a formant decomposition method to extract the evolution in time of the frequencies with the highest signal energy, which matches the resonance frequencies of the vocal tract. The entire vocal frequency spectrum was represented in the spatial dimension of the network's input, and the amplitude of the formants represents the evolution in time of the energy per frequency. 
The spatio-temporal features from the spoken digits were encoded and processed in a three-layer \gls{SNN} for the classification. The layers consist of a first encoding layer (L0) that encodes the amplitude of the formants into spike trains, a hidden layer (L1), and an output layer (L2) in which every neuron represents a keyword class. The keyword class was decoded from the most active neuron in layer L2 after training. 
We used a spoken digits dataset, applied speech decomposition tools to extract the formants, and compared the network performance with different neuron models in the hidden layer L1. All networks were trained to maximize the accuracy in the classification task according to supervised learning methods applied to \glspl{SNN}.
In the following sections, we explain the structure and preprocessing of the dataset, the comparison between network architectures, and the training methods for classifying spoken digits with \glspl{SNN}.

\noindent
\begin{minipage}{\linewidth}
\makebox[\linewidth]{
\includegraphics[keepaspectratio=true,scale=0.14]{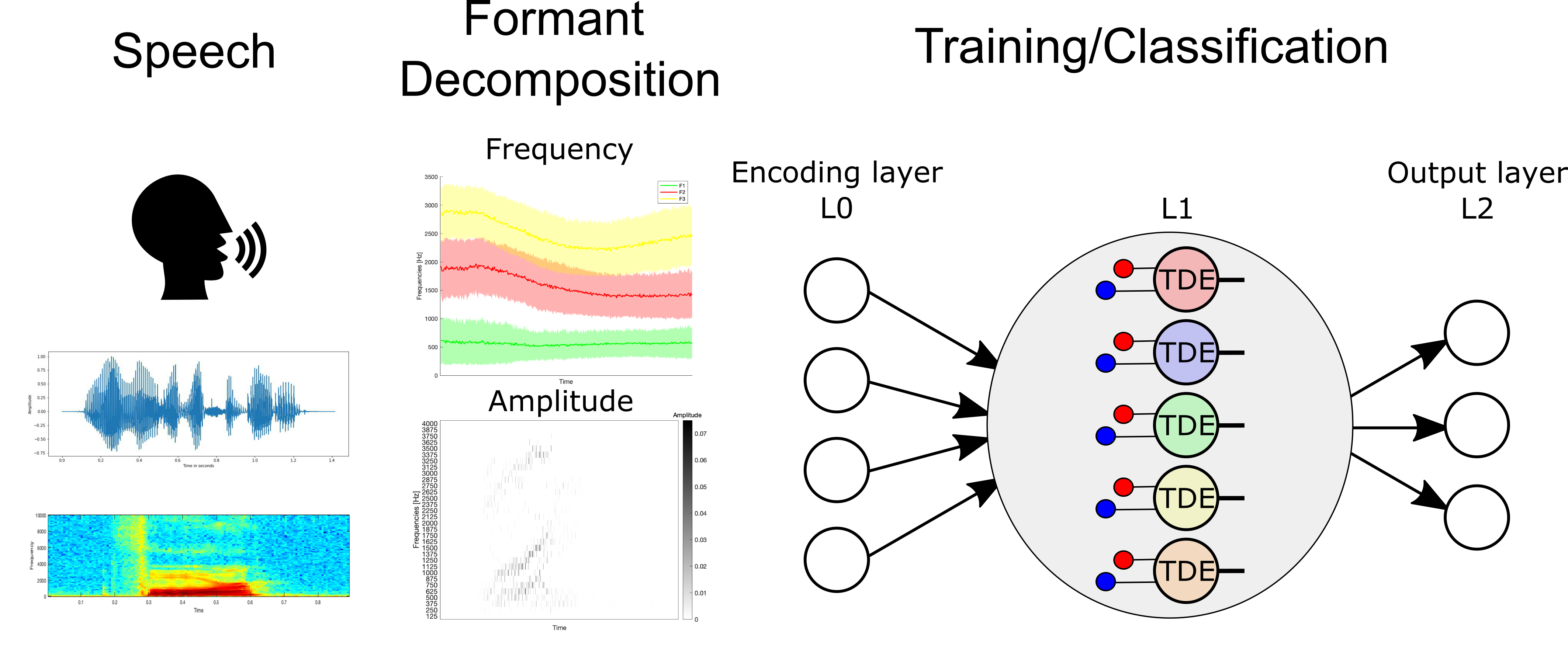}}
\captionof{figure}{System pipeline. From speech generation, we extract offline formant decomposition in frequency and the amplitude of the three formants with the biggest energy. Frequency is then represented as spatial information in the number of input neurons and amplitude of the formants is encoded in L0 into spikes. Network parameters are trained for performance in the classification task using backpropagation through time and moment estimation optimizer.}
\label{fig:method_system}
\end{minipage}

\subsection{Dataset and spike-based encoding}
\label{sec:encoding}
To test the performance of the \glspl{SNN} in keyword spotting, we used a popular speech recognition dataset, TIdigits~\citep{leonard1993tidigits}, containing speech utterances from 225 adult subjects from 11 classes (10 digits plus 'oh' the alternative utterance for zero). The dataset was sampled at 20 kHz for a total of 450 repetitions per class. The spatio-temporal sequences of words last approximately 400 ms and were presented at random initialisation times. We built the dataset with a fixed duration of 1500 ms, keeping this random vocalisation in place. Each of the keywords had been preprocessed into the main formants of the words as keyword features for the classification. 
The formants are defined in speech science as the acoustic resonances of the vocal tract and correspond with the main frequency components of the signal. Previous works used these features for word discrimination and speech recognition, where it is estimated that three formants are sufficient to discriminate vocal and semi-vocal sounds~\citep{Welling1998, Laszko2016}. 
We used the sinewave speech analysis toolbox in Matlab~\citep{sws2004} to extract the main three formants for each keyword. The formant frequencies have associated amplitudes that determine the energy of each formant. The temporal sequence of the amplitude at each frequency defines the spatio-temporal patterns that characterise each keyword class. 
We quantified the frequency spectrum into 32 equally spaced frequency bands, mapping the typical frequency range in human speech (0-4 kHz) into channels with a width of 125Hz. Each of these channels codifies the maximum amplitude of the three main formants into a temporal sequence of frequency amplitudes. Since the frequency channels have a frequency bandwidth, there is the possibility of a coincidence of 2 formants within the same input channel, which we solved by assigning the maximum possible amplitude to that frequency band.

The first layer of the classification network L0, composed of \gls{CuBa-LIF} neurons, converts the input sequence of amplitudes into spikes through the current of the neurons. Each of the 32 channels was connected to a single neuron for the spike conversion. Therefore, every cell in L0 represents a frequency band in the formant space of the keywords, creating a spatio-temporal spike code in which the network input space represents the frequency domain that evolves through time in the progression of the formants for each of the keywords.
Figure \ref{fig:method_formant} illustrates this conversion for 2 sample keywords of the words "oh" and "three". The left side of the figure represents the evolution in time of the frequency channels and their respective associated amplitude. Once the sequence of frequency amplitudes passes through the encoding layer of the network L0, the formants are transformed into spikes, as seen on the right side of the figure. 
This conversion is homogeneous for all frequencies with the same parameters for the neurons representing the frequency bands. The encoding layer L0 and the decoding layer L2 were fixed in terms of type, number, and cell parameters for all the networks compared in this work. The use of formant frequencies reduces the amount of information in the frequency domain that receives the encoding layer to the most relevant features in speech recognition, as seen in the visual distinction of the different keyword classes in the figure.

\noindent
\begin{minipage}{\linewidth}
\makebox[\linewidth]{
\includegraphics[keepaspectratio=true,scale=0.3]{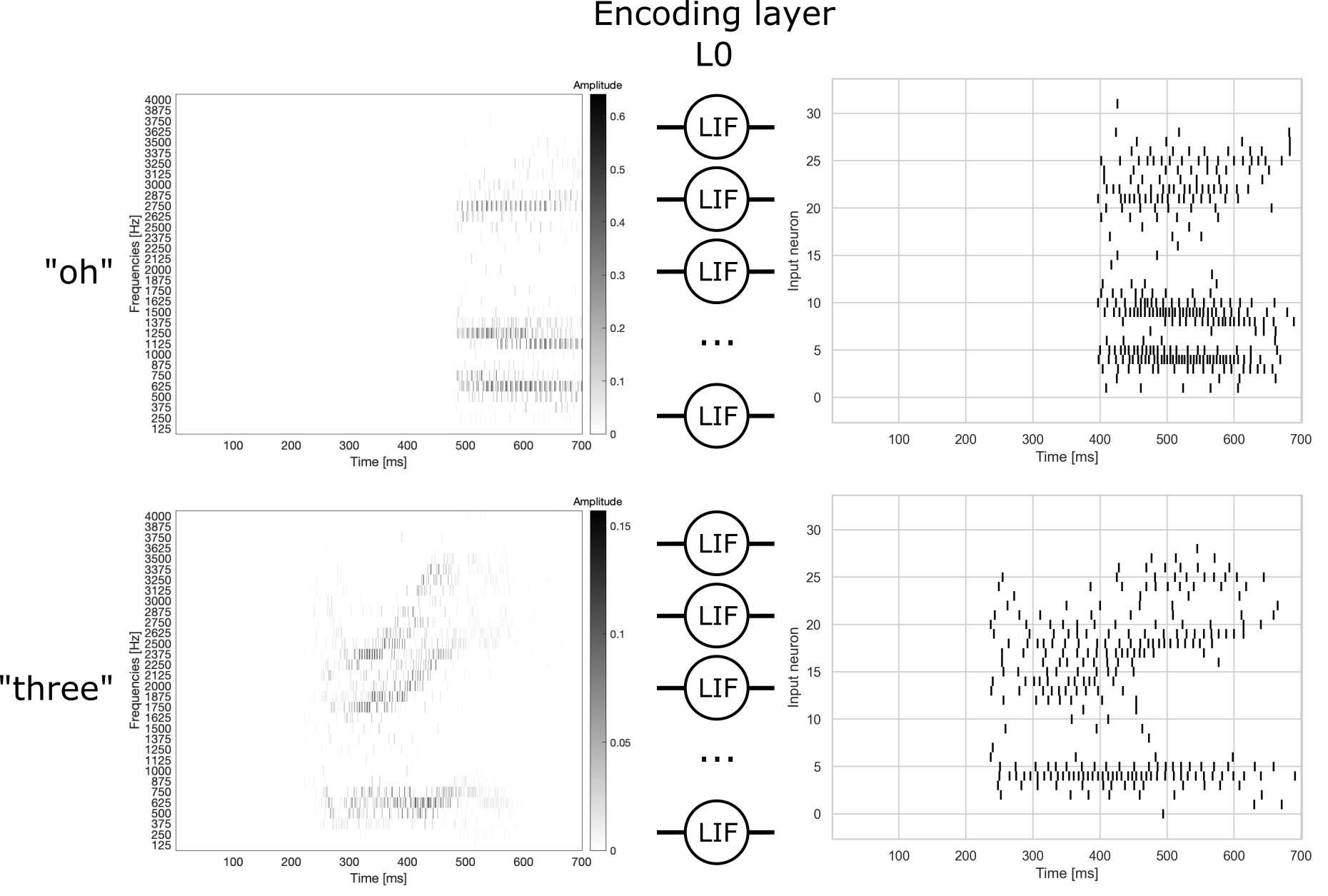}}
\captionof{figure}{Example of formants transformed into spikes for 2 sample keywords. Spoken keywords are decomposed in formant frequency bands and transformed into spikes in the encoding layer of the networks. Left side, sequence in time of frequency amplitude. Right side, the transformation of amplitude into spikes through the current of the \gls{CuBa-LIF} neuron.}
\label{fig:method_formant}
\end{minipage}

\subsection{Networks architectures}
\label{sec:cell_types}
We built different \glspl{SNN} based on the \gls{TDE} neuron model and the \gls{CuBa-LIF} neuron model, as seen in Figure \ref{fig:methods_arch}. While network layers L0 and L2 are formed with \gls{CuBa-LIF} neurons for all the architectures, our work shows how different models in the hidden layer L1 affect the performance, efficiency, and interpretability of the classification.
We conceptualize layer L0 as an encoding layer since the formant's amplitude from the speech preprocessing is converted into a train of spikes, assigning each of the neurons in L0 the encoding of a different frequency band. On the other end of the network, every neuron in layer L2 represents one keyword class in the classification problem. The class is selected based on the neuron with higher activity at its output. For all the networks in this comparative study, layer L1 was fully connected to layer L2, and the weight parameter ($W_2$) of this connection was trained for all the networks in this work. The network types benchmarked are described in the following text as (i) \textit{TDE network}, whose layer L1 is composed of \gls{TDE} neurons; (ii) \textit{LIFrec network} with \gls{CuBa-LIF} cells and recurrent connectivity in the L1 layer, (iii) and \textit{LIF network} that is composed of \gls{CuBa-LIF} cells with feedforward connectivity between layers. 

\noindent
\begin{minipage}{\linewidth}
\makebox[\linewidth]{
\includegraphics[keepaspectratio=true,scale=0.1]{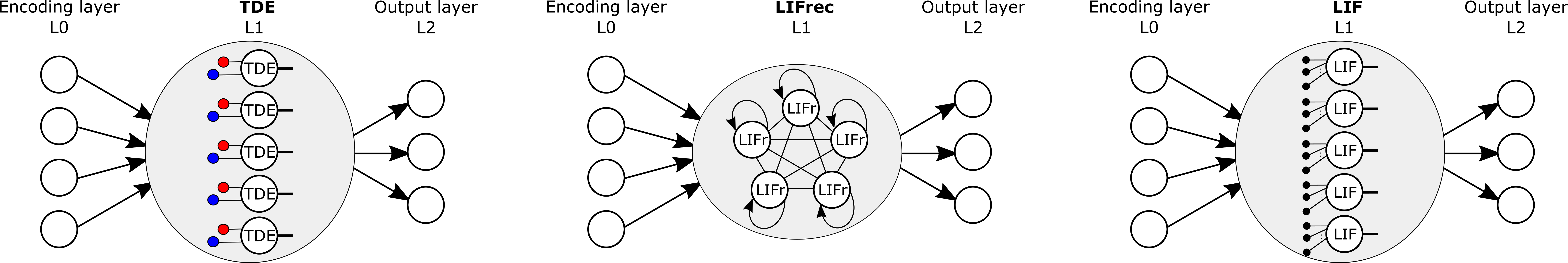}}
\captionof{figure}{Network architectures compared in the manuscript. All networks present the same number of neurons and the same type of model in layers L0 and L2. We balanced the number of synapses and a number of trained parameters for comparison.}
\label{fig:methods_arch}
\end{minipage}

\subsubsection{Current-Based Leaky Integrate-and-Fire (CuBa-LIF)}
The \gls{CuBa-LIF} model was adopted from previous work~\citep{cramer2020,zenke2021remarkable,Muller-Cleve2022}. 
This model, core of the comparative study, is described based on its membrane potential of the neuron $i$ in the layer $l$ with $U_i^{(l)}$ as,

\begin{equation}
\label{eq:snn1}
\tau_{mem} \frac{\mathrm{d}U_i^{(l)}}{\mathrm{d}t}=-(U_i^{(l)} - U_{rest}) + RI_i^{(l)}
\end{equation}
with $U_{rest}$ as membrane potential, $\tau_{mem}$ as the membrane time constant that defines the time dynamics of the cell, R as resistance, and $I_i^{(l)}$ the input current of the cell. Similarly, the $I_i^{(l)}$ is described as the combination of inputs from the previous layer $l-1$ accounting for the weights as: 
\begin{equation}
\label{eq:snn2}
    \tau_{syn} \frac{\mathrm{d}I_i}{\mathrm{d}t}= -I_i(t) + \sum_jW_{ij}S_j^{(l-1)}(t) + \sum_jV_{ij}S_j^{(l)}(t)
\end{equation}
with $\tau_{syn}$ the synaptic time decay of the input, $\sum_jW_{ij}S_j^{(l-1)}(t)$ describes the combination of the input contributions $S_j$ of the presynaptic neuron $j$ combined with the correspondent weight $W_{ij}$. Moreover, the contribution $\sum_jV_{ij}S_j^{(l)}(t)$ only applies when we use recurrency in our neuron model as what we described in the text as recurrent \gls{LIF} architecture or the abbreviation LIFrec.
\\
Discretizing the previous equations for the modelling into:
\begin{equation}
\label{eq:snn3_a}
    I_i^{(l)}(t)=\alpha I_i^{(l)}(t-1) + \sum_j W_{ij} \cdot S_j^{l-1}(t)
\end{equation}

\begin{equation}
\label{eq:snn3_b}
    U_i^{(l)}(t)=(\beta U_i^{(l)}(t-1) + I_i^{(l)}(t)) \cdot  (1.0 - U_{reset})
\end{equation}
with $\beta=\mathrm{exp}(\frac{-dt}{\tau_{mem}})$ as the voltage decay constant of the membrane, $\alpha=\mathrm{exp}(\frac{-dt}{\tau_{syn}})$ the current decay constant at synaptic state. We use for convenience an input resistance $R=1\Omega$ and $U_{reset}$ the voltage to reset the membrane after evoking a spike event.

\subsubsection{Time Difference Encoder (TDE)}
\label{subsec:methods:tde}
The \gls{TDE} is a processing cell that encodes the time differences between trains of spikes; this mechanism is hypothetically relevant for exploiting the temporal structure of the population code present in all sensory modalities in the nervous system~\citep{Panzeri2014}. While the first model was bio-inspired from the optical flow neural circuitry in the visual modality of biological organisms~\citep{Milde2015}, additional work has probed the potential use in a broader range of sensory modalities that involves vision, audition, touch and olfaction~\citep{Coath2014, Angelo2020, Schoepe2021, mastella2021hardware, Gutierrez-Galan2022}. 

The \gls{TDE} model consists of a pre-synaptic compartment with two inputs connected to a \gls{CuBa-LIF} neuron, as seen in Figure \ref{fig:methods_tde}. The spike rate in the neuron's output is inversely proportional to the time difference between the two inputs of the cell in a directional matter. The directionality plays an important role in the cell since the output trains are elicited only when facilitatory input receives a spike before the triggering input, and not the other way around. Hence, two different \gls{TDE} cells are potentially created from a pair of frequencies, one for each frequency combination. 
A spike in the facilitatory input (red trace) elicits a trace in the Gain compartment characterised by the time constant $\tau_g$. When a spike arrives at the trigger input (blue trace), the \gls{EPSC} of the cell is increased by the value of the gain at that time. The resulting \gls{EPSC} presents the same dynamics of the \gls{CuBa-LIF} neuron model in which a cell membrane integrates the current of the \gls{EPSC} trace to generate spikes in its output. The resulting equations are as follows: 

\begin{equation}
\label{eq:tde:dif}
        \tau_{syn} \frac{\mathrm{d}I_i}{\mathrm{d}t}= -I_i(t) + \sum_jG_{i}(t)S_{j,ithrig}(t)        
\end{equation}

\begin{equation}
\label{eq:tde2:dif}    
\tau_{gain} \frac{\mathrm{d}G_i}{\mathrm{d}t}= -G_i(t) + \sum_jW_{i,j}S_{j,ifac}(t)
\end{equation}

where compared with the \gls{CuBa-LIF} model, the current $I_i(t)$ has a time dependency in the weight $G_i(t)$ that multiplies the trigger connection. This synaptic compartment defined as $gain$ computes the spikes in the facilitatory connection with a leak governed by the time constant $\tau_g$. Similar to the previous discretization the resulting equations are as follows: 

\begin{equation}
\label{eq:tde}
    I_i(t)=\alpha I_i(t-1) + \sum_j G_i(t) \cdot S_{j,thrig}(t) 
\end{equation}

\begin{equation}
\label{eq:tde2}    
	  G_i(t)=\gamma G_i(t-1) + \sum_j W_{i,j} \cdot S_{j,fac}(t)	
\end{equation}

where $\gamma$ represents the exponential decay $\gamma=\mathrm{exp}(\frac{-dt}{\tau_{g}})$ with $\tau_{g}$ as an extra parameter of the cell which modulates the time effect of the correlation between the two inputs. $ W_{i,j}$ represents an additional weight assigned to the facilitatory spike that we keep at fixed value 1 in all cells. This simplification is made to exploit the temporal dynamics of the \gls{TDE} cell through the time constant instead of utilising the weight matrix ($W_1$) which is assigned to the frequency input. Therefore, the \gls{TDE} aggregates a pre-synaptic compartment in which the gain of the trigger connection is modulated by the temporal trace of the facilitatory input, as represented in the figure. $\tau_g$ governs the leakage of the gain compartment assigned to the facilitatory input. It is a key parameter of the cell since it controls the effect of the gain and the effective time distance between the spike trains between facilitatory and triggering inputs. This was trained for each of the cells in L1 in \textit{TDE network} since every cell has a different pair of connections from the encoding layer L0. For instance, a low value of the time constant $\tau_g$ filters long-time differences between spikes in the cell's input. While in a cell with a high value of $\tau_g$, a long time distance between spikes at its input elicits a neural response in the synaptic trace. Training this parameter, we adjusted the time constant dynamic for each pair of frequencies in the frequency spectrum.

\noindent
\begin{minipage}{\linewidth}
\makebox[\linewidth]{
\includegraphics[keepaspectratio=true,scale=0.4]{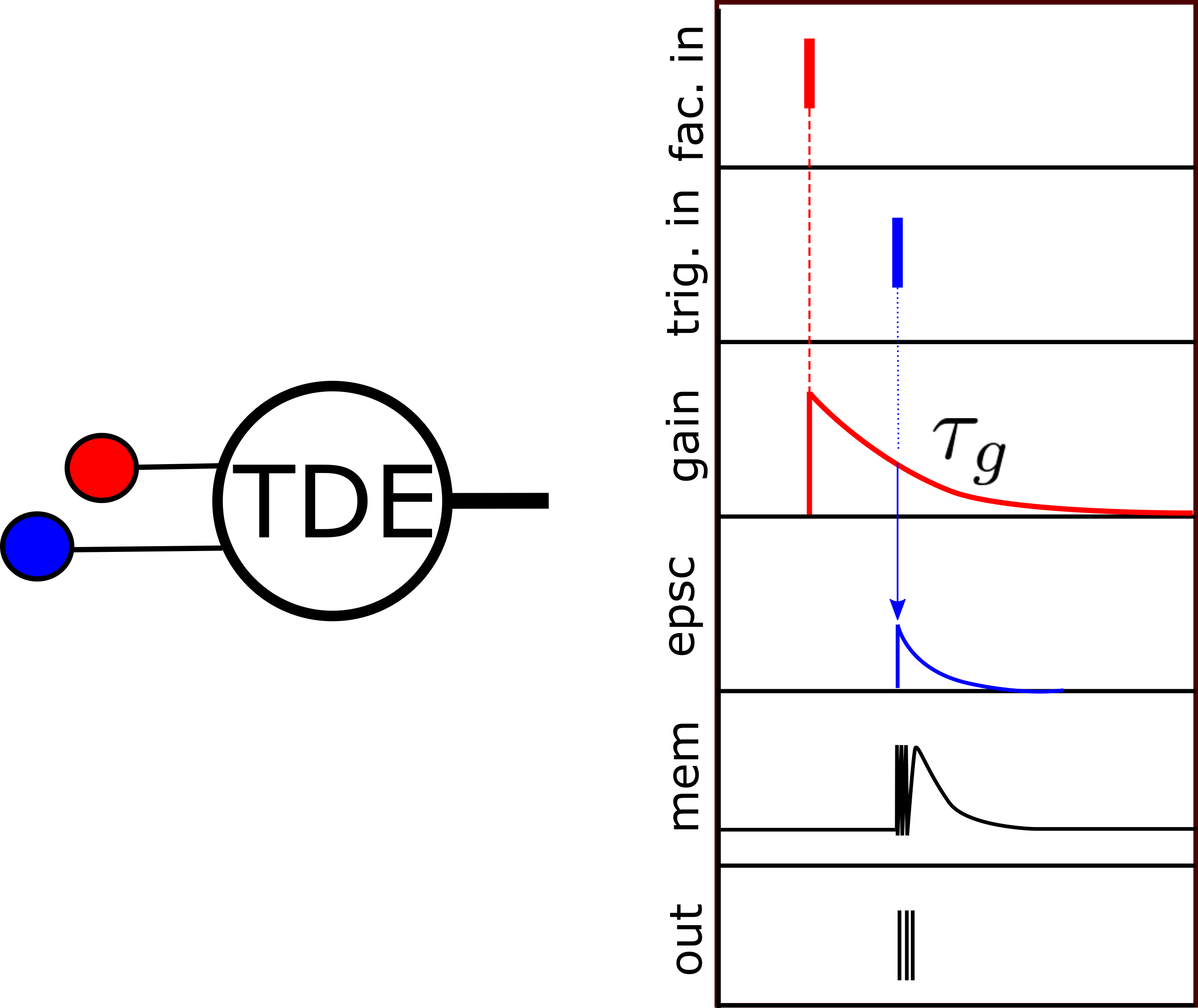}}
\captionof{figure}{Time Difference Encoder (TDE) in the presence of input spikes. $\tau_g$ parameter governs the effect of the distance in time between facilitatory and trigger spikes. Adapted from~\citep{Angelo2020}.}
\label{fig:methods_tde}
\end{minipage}

\subsection{Offline training with back-propagation through time}
We used \glspl{SNN} for keyword classification and optimized the network parameters with supervised learning methods. We used \gls{BPTT} with surrogate gradient descent that uses a surrogate function in the backward propagation of the network loss $\mathcal{L}$, bypassing the spike discontinuity that represents an issue with applying the true gradients. This method has been proven to be highly effective in training \glspl{SNN}~\citep{paszke2017automatic,neftci2019surrogate,zenke2021remarkable}. Following previous works, we exploited the differentiation capabilities by overloading the derivative of the spiking non-linearity with a differentiable fast sigmoid function with $\lambda$ as the scale parameter of the slope as:

\begin{equation}
\label{eq:snnSig}
    \sigma(U_i^{(l)})=\frac{U_i^{(l)}}{1 + \lambda |U_i^{(l)}|}
\end{equation}
In order to apply the back-propagation method to the train of spikes, we first used the cross entropy loss function over the active number of spikes in the output layer (L2 in the network architecture) whose each individual cell represents a keyword class. Formalizing the function for a given batch number ($N_b$) and number of classes ($N_c$) for the loss function as:

\begin{equation}
\label{eq:loss}
    \mathcal{L}=- \frac{1}{N_{b}} \sum_{s=1}^{N_{b}} \mathds{1}(i=y_s) \cdot \mathrm{log}\left\{\frac{\mathrm{exp}\bigg(\sum_{n=1}^T S_i^{(l)} [n]\bigg)}{\sum_{i=1}^{N_{c}} \mathrm{exp}\bigg(\sum_{n=1}^T S_i^{(l)} [n]\bigg)} \right\}
\end{equation}

then we propagated the loss function to the parameters of the network as: 
\begin{equation}
\label{eq:sn4}
    W_{ij} \leftarrow W_{ij} - \eta \frac{\partial \mathcal{L}}{\partial W_{ij}} \quad \textrm{,} \quad \tau_{g} \leftarrow \tau_{g} - \rho \frac{\partial \mathcal{L}}{\partial \tau_{g}}
\end{equation}

where $W_{ij}$ describes the training of the weights and the $\tau_{g}$ the parameter of the \gls{TDE} cells for the network as described in Section \ref{subsec:methods:tde}. The parameters $\eta$ and $\rho$ define the learning rate. Then, we used an \gls{ADAM} optimizer algorithm to update the parameters based on the gradient functions with dropout and weight decay.
These two methods have became standard methods in neural networks to improve the speed in the generalization of the training and avoid over-fitting~\citep{srivastava2013, smith2018}. The rest of the hyperparameters of the network were empirically selected based on performance and are shown in Table \ref{tab:hyperparam}. The results found here correspond to cross-validation testing results after training each network individually.

\begin{table}[ht]
    \caption{Description of the hyperparameters used in the supervised learning procedure. Hyperparameters tuned from exploration based on previous works in training \glspl{SNN} from spatio-temporal patterns with biological constraints~\citep{Muller-Cleve2022}.}
    \label{tab:hyperparam}
    \centering
    \begin{scriptsize}
        \begin{tabular}{>{\centering}m{0.2\linewidth}>{\centering\arraybackslash}m{0.4\linewidth}>{\centering\arraybackslash}m{0.2\linewidth}}
	        \textbf{Hyperparameter} & \textbf{Description} & \textbf{Value} \\
	        \toprule
	        \textbf{scale}, $\lambda$ & Steepness of surrogate gradient & 5\\
	        \midrule
	        \textbf{time\_bin\_size} & Time binning of the encoded input & 0.015 [s]\\
	        \midrule	        
          \textbf{sim\_step\_size}, $dt$ & Time between steps of the simulation & 0.015 [s]\\
	        \midrule	        
	        \textbf{tau\_mem}, $\tau_{mem}$ & Decay time constant of the membrane & 0.002 [s]\\
	        \midrule
	        \textbf{tau\_syn}, $\tau_{syn}$ & Decay time constant of the synapse & 0.008 [s]\\
          \midrule	      	     
	        \textbf{learning\_rate}, $\mu,\rho$ & learning rate &  0.0015\\
	        \midrule	      	     
	        \textbf{weight\_dec} & weight decay of the optimizer in every step &  0.0001\\
	        \midrule
	        \textbf{p\_drop} & Probability of training dropout & 0.1\\
	           \hline
        \end{tabular}
    \end{scriptsize}
\end{table}

\subsection{Comparison of the performance of different network models}
\label{subsec:conn}
We trained three networks and compared the results' performance, efficiency, and interpretability. The networks are represented in Figure \ref{fig:methods_arch} as TDE, LIFrec, and LIF networks. All networks explored consist of three layers. Input layer L0 is defined as the encoding layer since it converts the energy of the formant frequencies into spikes and consists of 32 input neurons as described in Section \ref{sec:encoding}. The hidden layer L1 consists of different neuron models based on the description in Section \ref{sec:cell_types}. The output layer L2 is a fully connected layer with a parameter to optimise defined as $W_2$. The keyword class was decoded as the cell with higher spike activity in L2, which is formed by 11 cells, one cell for each of the classes in the dataset. 

To compare the networks, we balanced the number of cells in the hidden layer L1 based on the total number of synaptic connections. We first calculated the number of synaptic connections in the TDE network, which utilises all combinations of pairs in L0 in the biggest possible network. Then, we balanced the number of cells needed in the LIF and LIFrec networks to account for the same number of synaptic operations given the fixed number of cells in the L0 and L2 layers. The number of synaptic connections for each of the network types is described in function of the number of cells in L1 ($N_{L1}$) as:

\begin{equation}
\label{eq:connections}
\begin{aligned}
    &TDE \quad Conn = N_{L1} \cdot 2 + N_{L1} \cdot N_{L2} \\
    &LIFrec \quad Conn = N_{L1} \cdot N_{L0} + N_{L1} \cdot N_{L1} + N_{L1} \cdot N_{L2} \\
    &LIF \quad Conn = N_{L1} \cdot N_{L0} + N_{L1} \cdot N_{L2} \\
\end{aligned}
\end{equation}


The same number of connections means that the three networks have the same synaptic weight capacity, which serves as a proxy of the memory footprint used by a hardware implementation. The number of cells was rebalanced for each of the \gls{TDE} pruned networks that leads the comparative study. 


We compared the networks in terms of keyword classification performance, computational complexity, and interpretability of the trained network. Measurements of complexity as the number of synaptic operations needed for a decoded utterance are commonly used as a proxy for energy consumption~\citep{caccavella_etal23,yik2023}. 
When comparing performance in keyword spotting, we train each network type according to its defined parameters. TDE network parameters are $\tau_g$ and $W_2$, LIFrec network parameters defined as $W_1,W_{11},W_2$, and LIF network learning $W_1,W_2$ parameters respectively. For the training procedure, we randomly divided the dataset between train and test, keeping the same number of samples per class. During each epoch, the training loop iterates through all the training datasets, optimizing the parameters to minimize the loss function and calculating the classification accuracy of the non-trained test dataset over a fixed number of epochs. The reported performance in the accuracy comparison results from averaging the 25 best testing accuracy results across the training epochs. This method reduces the random variability of the testing accuracy. 

Regarding efficiency, the energy expenditure was approximated by the sparsity level in decoding all spatio-temporal sequences. Targeting a neuromorphic implementation of \glspl{SNN}, we considered two event-driven asynchronous computations at the level of the neuron cell that increases the dynamic energy consumption of the system following previous analytical works~\citep{lemaire2022}: (1) the operation of integrating an input spike (i.e., add to the membrane potential), and (2) the operation of firing an output spike (i.e., reset of the membrane potential). We calculated the number of \gls{SynOps} as the sum of all neuron $j$  for each layer as described in the following Equation \ref{eq:synops}:

\begin{equation}
	 SynOps = \sum_j S_{in,j} + \sum_j S_{out,j}
\label{eq:synops}
\end{equation}

Concerning the interpretability of the results, in every comparative network, the connectivity matrix $W_2$ represents the combination of features in L1 that contributes to maximizing the spike rate of each of the classes in the output layer L2. Compared with the other networks, where the connectivity matrix was not fixed in selectivity nor quantity (parameter $W_1$), the TDE network presents a fixed connectivity in which every \gls{TDE} cell represents a different pair of frequencies from the input data in layer L0. Therefore, interpreting the obtained results in $W_2$ in the TDE network, we quantify the contribution of each pair of frequencies in the input layer to each of the keyword classes in L2 and the time constant that maximizes the classification $\tau_g$.


\section{Results}
\subsection{Time pattern versus rate}
The resulting spiking data from converting formants into spikes presents a question about how we can exploit this information for the classification task. Several works in the audio modality have already highlighted that speech, by its nature, conveys a large amount of information in the time domain~\citep{Kayser2010}. Since \glspl{SNN} have been known to exploit the information about time more efficiently~\citep{Yarga2022}, we investigated these findings in the chosen dataset. For that purpose, we used information theory to quantify and compare the amount of information carried in spiking rate, removing the time information out of the signal, and about spiking time, measuring the time precision of the elicited spike patterns after encoding. This methodology has been previously used to calculate the precision of the spike trains~\citep{Kayser2009}. These measurements are termed $I_{rate}$ and $I_{pattern}$ respectively and are described in the left panel of Figure \ref{fig:results_info} along with the results of the metric applied to the dataset. In the figure, we combined as mean and standard deviation the measurements about $I_{rate}$ and $I_{pattern}$ from each of the 32 neurons that encode the information in all the frequency bands of the formants. In the figure, $\Delta t$ represents the time bin which defines the precision of the time pattern code. For this analysis, we calculated the spike patterns from the first spike for a total duration of 400 ms. The results in the right side of Figure \ref{fig:results_info} show that the information about spike timing is even bigger than about the rate when we use a big time window, and it is equivalent in the range of 15 to 20 ms. Hence, this demonstrates that the information about time is relevant to the classification task.

\noindent
\begin{minipage}{\linewidth}
\makebox[\linewidth]{
\includegraphics[keepaspectratio=true,scale=0.4]{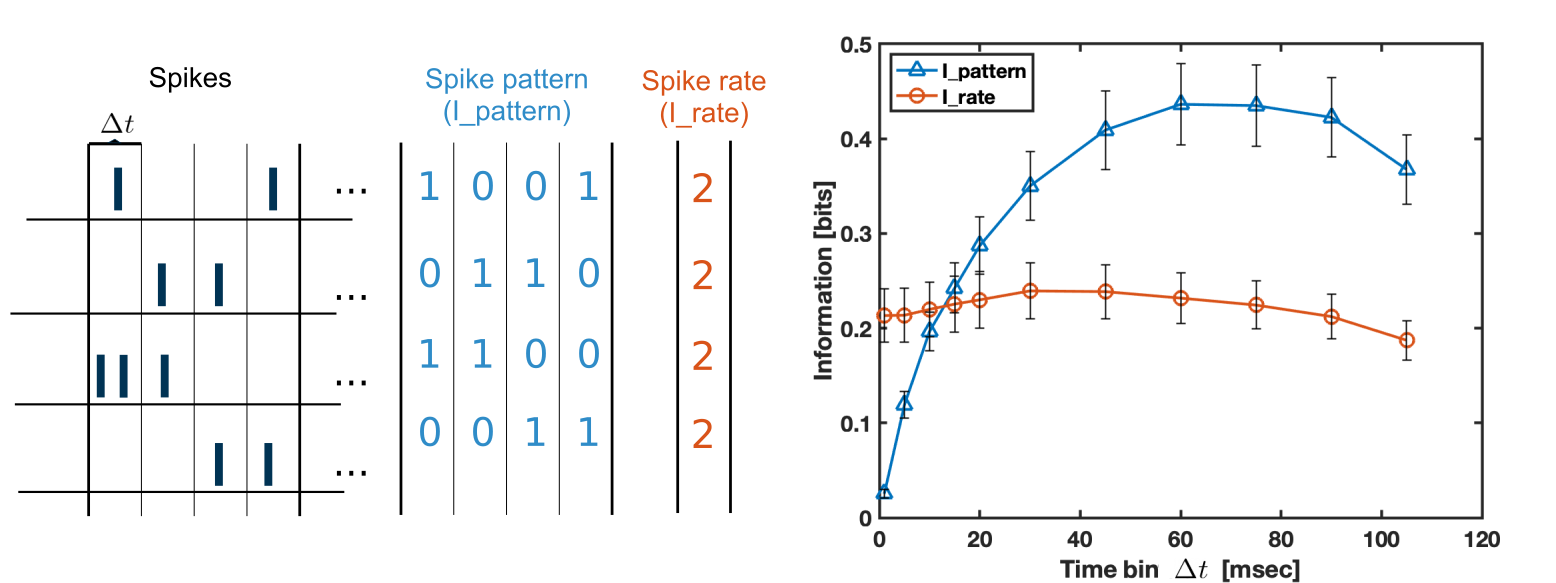}}
\captionof{figure}{Information about the spike pattern and spike rate of the encoded spikes in the dataset. The left figure represents the methodology for calculating I\_pattern versus I\_rate from the spike trains. The right figure shows the amount of information about the spiking rate against the spiking pattern in time of spikes in the encoded dataset. The mean and standard deviation of the figure show the variation for each of the individual frequency channels encoded, which served as input to the networks in this study.}
\label{fig:results_info} 
\end{minipage}

\subsection{Data-driven TDE-based network architecture} \label{subsec:crosscorr}
In the proposed network with \gls{TDE} cells, every cell connects two inputs that combine two frequency bands in the proposed data given the previously studied \gls{TDE} model (Sec. \ref{subsec:methods:tde}). Given that every \gls{TDE} cell is defined for its determined input, this cell model can be tuned for the expected input dataset of the task, resizing the network with the expected number of \gls{TDE} cells and guiding an estimate of the parameter initialisation for the network's training. In comparison, typical architectures with \gls{LIF} and recurrent \gls{LIF} cells present full connectivity and random initialisation. In this type of network, the complexity and amount of hardware resources are selected through the number of cells per layer. Therefore, by building architectures with \gls{TDE} cells, we can customize and gain insights into the estimated number of cells analyzing the expected data input. This analysis yields the rank of the most activated gls{TDE} cells that expect to detect features for the given input.
Taking a data-driven approach, we measured the unbiased cross-correlation values to every frequency pair in the spike encoding of the formants from L0. With this process, we identified the highest correlated pairs of frequencies connected to the cells in our network. Since the \gls{TDE} cell behaviour maximises its output as a coincidence detector, we ranked the \gls{TDE} cells according to their cross-correlation highest values. Figure \ref{fig:results_crosscorr} presents the results of this analysis. In the left column, the top heatmap represents the average within the word class and the maximum across classes of the cross-correlation measurement. The bottom heatmap shows the average of the lags as they correspond to the maximum cross-correlation value expressed in the top figure. This value served as an initialisation of the time constant $\tau_g$ that defines the pre-synaptic connection of each \gls{TDE} cell. 
Given the cross-correlation value, we ranked the \gls{TDE} cells in layer L1 and redimension the network prioritising the \gls{TDE} cells with higher cross-correlation values. In the top right of Figure \ref{fig:results_crosscorr}, we show a histogram of every TDE cell with its associated cross-correlation value. Based on the cross-correlation ranked list, we selected the number of cells pruning the lowest cross-correlated values of its prospective input; the levels we chose to prune the network are represented with red lines in the figure. 
We trained each of the networks with the selected cells from this analysis and measured the corresponding test accuracy. In the bottom right of Figure \ref{fig:results_crosscorr}, we represent the informed pruning of the network based on the values of the analysis with a random pruning. The results output better performance at all pruning levels with the informed condition. A more compact view of the properties of these networks (number of cells, connections, and accuracy) can be seen in Table \ref{tab:cross-corr}. This initial dataset analysis reduced 45.5$\%$ in the number of neurons and connections with only a penalty performance of 1.22$\%$ for the network with 540 \gls{TDE} cells, as seen in the table. Given its efficiency, we used this network size in the following sections for a detailed comparison between network types.

\noindent
\begin{minipage}{\linewidth}
\makebox[\linewidth]{
\includegraphics[keepaspectratio=true,scale=0.45]{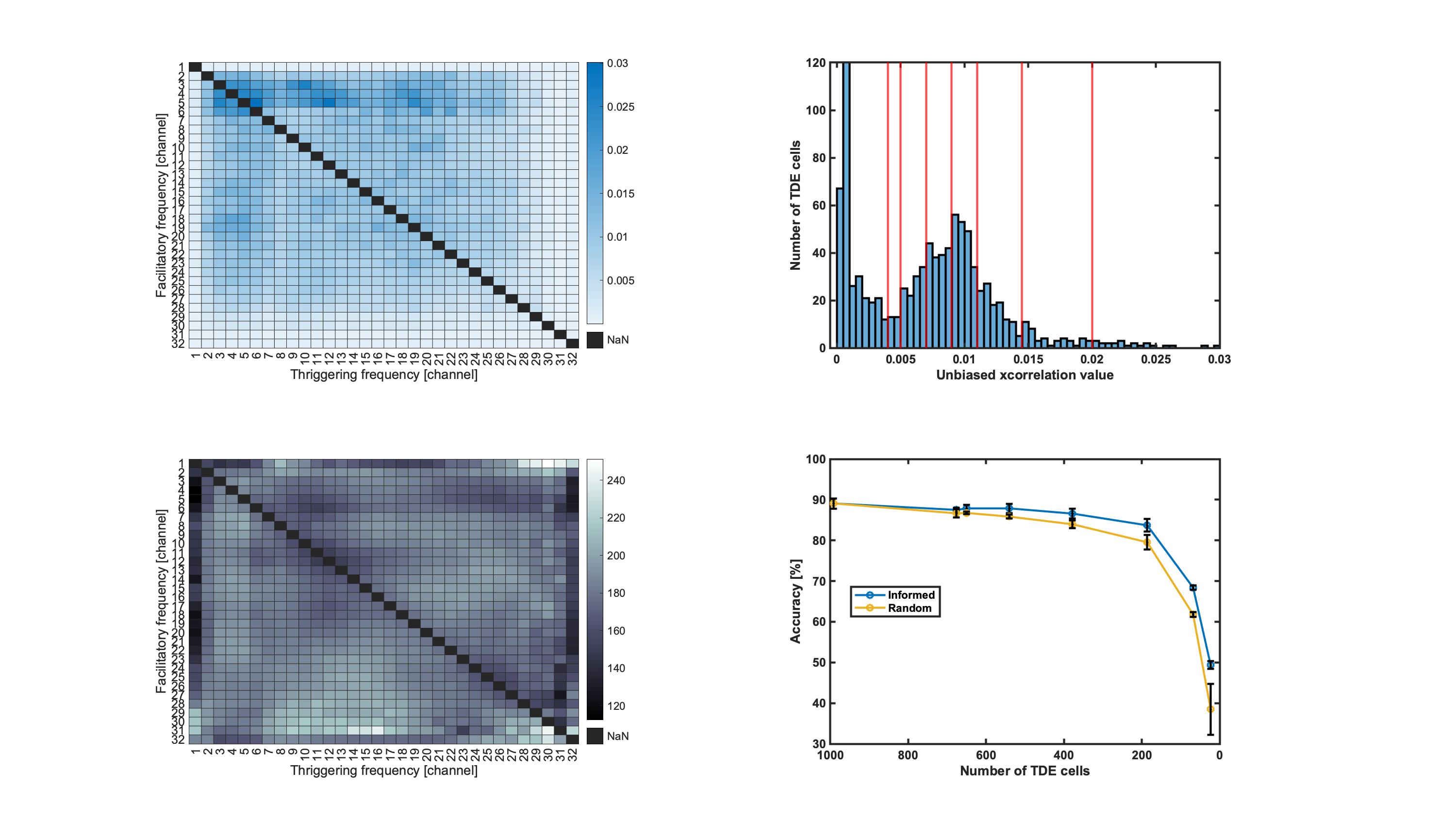}}
\captionof{figure}{Cross-Correlation values of pairs of frequencies and \gls{TDE} cell selection with accuracy results. Top-left: unbiased cross-correlation values for each frequency pair in the dataset are calculated as the mean within the class and maximum value across classes. Bottom-left: average lag between classes, which outputs the maximum value of cross-correlation. Top-right: histogram of all possible combinations of \gls{TDE} cells and associated cross-correlation values of spike input. Red lines defined the pruning levels selected. Bottom-right: comparative figure between data-driven informed pruning and random pruning. Accuracies associated with the training of networks at specified pruning levels and correspondent number of \gls{TDE} cells.}
\label{fig:results_crosscorr}
\end{minipage}

\begin{table}[ht]
\centering
\caption{Table of cross-correlation values for frequency pairs, an equivalent number of \gls{TDE} cells in the network, and comparison of the pruning of the network between the cross-correlation approach (x-corr) and random pruning (random) in terms of network performance for the \gls{TDE} architecture. The Cross-Corr column corresponds to the minimum cross-correlation value of the frequency pairs that correspond to the respective number of \gls{TDE} cells in the following column (\# TDE). These values are also represented in Fig.\ref{fig:results_crosscorr}. Informed by the dataset, we efficiently selected the \gls{TDE} cells to prune. Data-driven pruning of the \gls{TDE} cells facilitates the exploration of the trade-off between the resources for a hardware implementation and the performance. The accuracy values were calculated as the mean and standard deviation of 3 independent learning iterations. The calculated test accuracy is the resultant of the average of the top 25 best test accuracies after training using Backpropagation Through Time (BPTT).}
\label{tab:cross-corr}
\begin{tabular}{cccc} 
\multicolumn{1}{c}{\textbf{\begin{tabular}[c]{@{}c@{}}Cross-Corr \end{tabular}}} & \multicolumn{1}{c}{\textbf{\begin{tabular}[c]{@{}c@{}}\# TDE cells\end{tabular}}} & \multicolumn{1}{c}{\textbf{\begin{tabular}[c]{@{}c@{}}TDE Accs (std)\\ x-corr \end{tabular}}} & \multicolumn{1}{c}{\textbf{\begin{tabular}[c]{@{}c@{}}TDE Accs (std)\\ random \end{tabular}}}  \\ \hline
all    & 992 & 89.08 (1.25) & 89.08 (1.25)\\ \hline
0.004  & 676 & 87.49 (0.47) & 86.64 (0.97) \\ \hline
0.005  & 650 & 87.86 (0.88) & 86.64 (0.27)\\ \hline
0.007  & 540 & 87.86 (1.12) & 85.83 (0.46)\\ \hline
0.009  & 378 & 86.58 (1.22) & 83.95 (0.92)\\ \hline
0.011  & 186 & 83.7 (1.55)  & 79.53 (1.77)\\ \hline
0.0145 & 68  & 68.36 (0.54) & 61.77 (0.63) \\ \hline
0.02   & 23  & 49.35 (0.98) & 38.47 (6.30) \\ \hline
\end{tabular}
\end{table}

\subsection{Classification accuracy and inference efficiency}
For the following comparative study, we chose to prune the TDE network at different network size based on the cross-correlation analysis from the data as described in Section \ref{subsec:crosscorr}. We selected three network sizes in the TDE architecture (992, 540, and 186 cells). We balanced the number of neurons in the LIFrec and LIF network based on the number of connections of the network as a network complexity measurement (Section \ref{subsec:conn}).
Balancing the total number of connections instead of comparing the networks based on the number of cells allows us to better approximate the amount of resources that these networks allocate in a hardware implementation. Furthermore, it also approximates the complexity between architectures in terms of the amount of synaptic weights and trainable parameters. Table \ref{tab:network-comparative} compares all networks explored regarding classification accuracy. In the obtained results, the recurrent LIF (LIFrec) network presents the highest accuracy value with 91\% compared to the 89\% of the TDE architecture. These values correspond to the biggest network size tested, corresponding to all the dataset's frequency pair combinations. While all the networks explored decrease the accuracy when reducing the number of cells in layer L1, the feedforward LIF network reaches its maximum capability in around 160 neurons in L1. 
We explored further comparisons with the first pruned TDE network proposed since this network offers a positive tradeoff between smaller network sizes and a non-significant decrease in classification performance. Figure \ref{fig:results_spikes} and Table \ref{tab:acc_trans} present the comparative results between the three networks regarding test accuracy, the total number of elicited spikes, and the number of synaptic operations performed. The values for each network were calculated after the training phase. All the spikes and synaptic operations metrics were averaged per presented keyword out of the full dataset.
Synaptic operations were calculated as the total operations computed in the cells; we calculated the computational events in asynchronous hardware implementations, which are the ones elicited by a spike in the input of a cell and the operation to elicit a spike in their output. This metric combines the activation sparsity of the network with the connectivity between layers as a single metric of the computational cost in the network~\citep{yik2023}.

We found that, although the TDE network is, on average, ~2\% lower in accuracy, the number of spikes and the total synaptic operations in the networks are considerably lower for the TDE architecture. In comparison, the TDE network generates ~53\% fewer spikes and computes ~92\% less synaptic operations along the network against the highest performance of the LIFrec network. This result is of paramount importance because of the direct impact of the synaptic operations on the energy consumption of the neuromorphic computing system. 
When comparing different layers of synaptic operations, we observe how the considerable number of spikes in the LIFrec network's L1 not only elicits an increase in the number of synaptic operations in L1 (the layer where recurrent cells are) but also in the following layer, L2, which is formed by the same number of LIF cells in all three architectures of the comparative.


\begin{minipage}{\textwidth}
\begin{minipage}[b]{1\textwidth}
\centering
\includegraphics[keepaspectratio=true,scale=0.55]{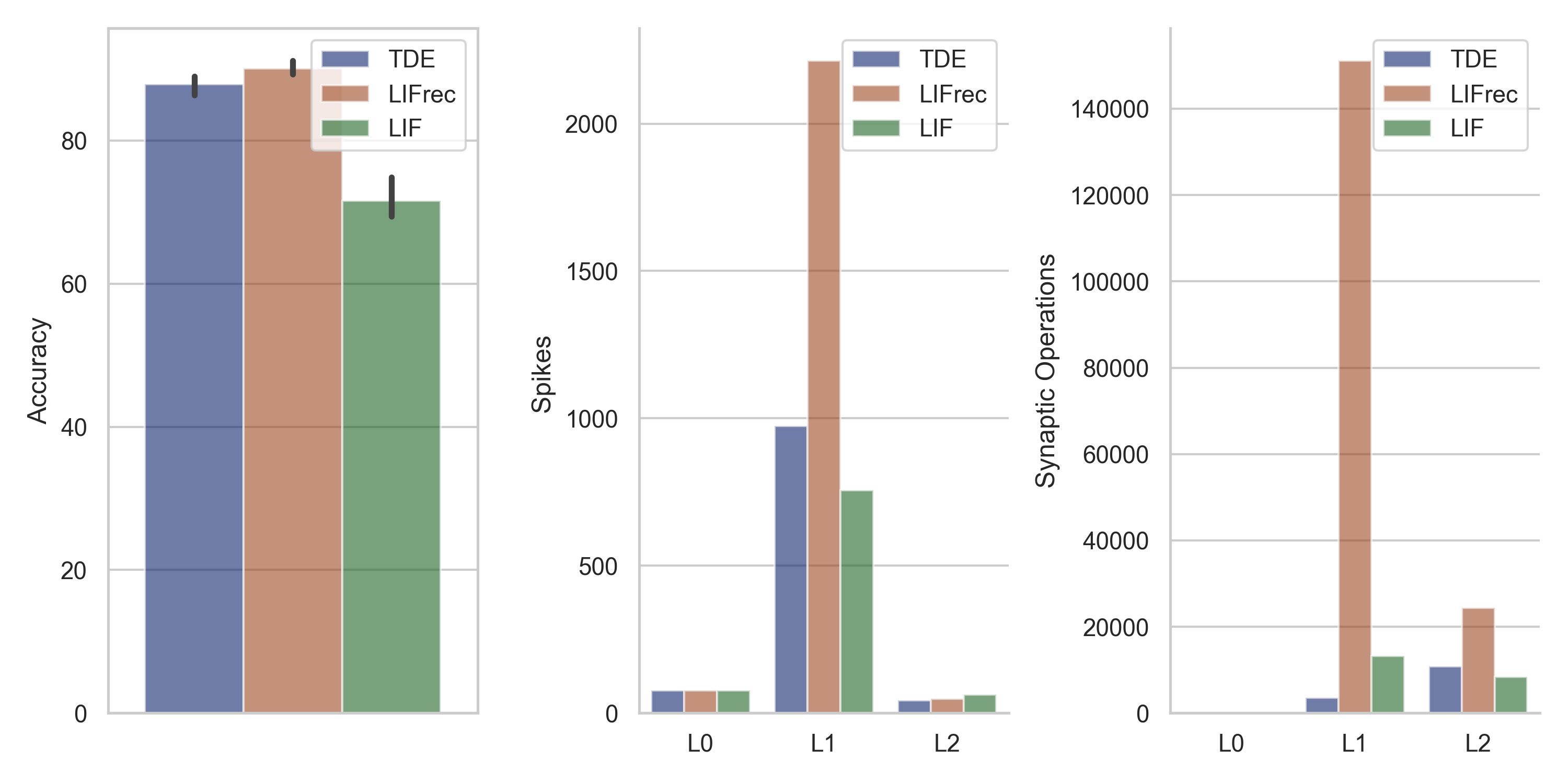}
\captionof{figure}{Comparison of spikes per layer and synaptic operations in the presentation of the full dataset. Spikes at each layer are calculated as the total number of evoked spikes at each layer for the whole dataset. Spikes in the middle subfigure are defined as the number of spikes in the output of all cells in their respective layer. Synaptic operations account for the connectivity of each of the layers, and every spike evokes an operation in event-based computation architectures. For instance, spikes in the output of L0 are computed in every single neuron of L1 for the LIF network since it is fully connected, but only a fraction of that is computed in a single cell of L1 for the TDE network.}
\label{fig:results_spikes}
\end{minipage}  
\vfill  
\begin{minipage}[b]{1\textwidth}
\centering

\begin{tabular}{llllllll}
 &  & \multicolumn{2}{l}{\textbf{TDE}} & \multicolumn{2}{l}{\textbf{LIFrec}} & \multicolumn{2}{l}{\textbf{LIF}} \\ \hline 
\multicolumn{2}{l}{\textbf{Accuracy (std)}} & \multicolumn{2}{c}{87.86 (1.12)} & \multicolumn{2}{c}{90.04 (0.8)} & \multicolumn{2}{c}{71.59 (2.72)}  \\ \hline 
\multicolumn{2}{l}{\textbf{\# cells L1}} & \multicolumn{2}{c}{540} & \multicolumn{2}{c}{65} & \multicolumn{2}{c}{163} \\ \hline
\multicolumn{2}{l}{\textbf{Synaptic connections}} & \multicolumn{2}{c}{7020} & \multicolumn{2}{c}{7020} & \multicolumn{2}{c}{7009} \\ \hline
\multicolumn{2}{l}{\textbf{Trained parameters}} & \multicolumn{2}{c}{6480} & \multicolumn{2}{c}{7020} & \multicolumn{2}{c}{7009} \\ \hline 
\multicolumn{2}{l}{\textbf{Spikes}} & \multicolumn{1}{c}{\#} & \multicolumn{1}{c}{Diff } & \multicolumn{1}{c}{\#} & \multicolumn{1}{c}{} & \multicolumn{1}{c}{\#} & \multicolumn{1}{c}{Diff} \\ \hline
\textbf{Layer} & L0 & 77 & 0.00\% & 77 &  & 77 & 0.00\% \\
\textbf{} & L1 & 975 & -55.95\% & 2213 &  & 757 & -65.78\% \\
\textbf{} & L2 & 43 & -12.37\% & 49 &  & 63 & 29.52\% \\
&Total & 1095 & -53.19\% & 2339 &  & 898 & -61.62\% \\ \hline
\multicolumn{2}{l}{\textbf{Synaptic Operations}} & \multicolumn{1}{c}{\#} & \multicolumn{1}{c}{Diff} & \multicolumn{1}{c}{\#} & \multicolumn{1}{c}{} & \multicolumn{1}{c}{\#} & \multicolumn{1}{c}{Diff} \\ \hline
\textbf{Layer} & L0 & 77 & 0.00\% & 77 &   & 77 & 0.00\% \\
 & L1 & 3572 & -97.64\% & 151076 &  & 13299 & -91.20\% \\
 & L2 & 10768 & -55.86\% & 24395 &  & 8394 & -65.59\% \\
&Total & 14417 & -91.79\% & 175548 &   & 21771 & -87.60\% \\ \hline
\end{tabular}
\captionof{table}{Comparative table of spikes and synaptic operations averaged per keyword over the whole dataset. Spikes and synaptic operations are calculated after training procedures. \# column presents the number, \textit{Diff} column describes the percent difference between the LIFrec network (the highest performant reference). Values calculated per layer and total. Networks are balanced based on synaptic connectivity.}
\label{tab:acc_trans} 
\end{minipage}
\end{minipage}
\input{tab/tab_comparison1}

\subsection{Training data efficiency}
The performance of neural networks (\gls{ANN} and \gls{SNN}) trained with ground truth (i.e., supervised learning) using backpropagation is highly sensitive to the size of the dataset, being the number of parameters in the network a well-known factor that correlates with the size of the required dataset to train the network~\citep{soekhoe2016,alwosheel2018}. 
Given the dataset presented for the task, we compared the training efficiency and dataset size sensitivity of the networks with the highest performance in the previous comparative, the TDE and the recurrent LIF (LIFrec) networks. For that purpose, we reduced the size of the training dataset keeping the test dataset at the same size to examine the inference power of the network against a smaller number of samples per class. The comparative results are shown in Figure \ref{fig:results_traineff}. From the figure, we appreciate a bigger decreasing slope for the LIFrec than for the TDE architecture, thus expressing a higher sensitivity to the decrease in the training size. The TDE architecture shows no statistically significant change in the performance with a reduction of the training dataset from the 100$\%$ available to the 75$\%$ while LIFrec shows a statistically significant decrease. The decrease in performance is generally more prominent for the LIFrec architecture, showing that TDE architecture, in comparison, is more robust and efficient in the presence of smaller datasets and a smaller number of samples per class. The first distinction between these networks is the number of trained parameters (7\% less parameters in the TDE network). However, the fixed connectivity of the TDE cells may play a more significant role in training efficiency. While the LIFrec network builds the feature map in the training stage, the TDE network enforces this connection between input and L1 by design, improving the training capabilities of the network. 

\noindent
\begin{minipage}{\linewidth}
\makebox[\linewidth]{
\includegraphics[keepaspectratio=true,scale=0.7]{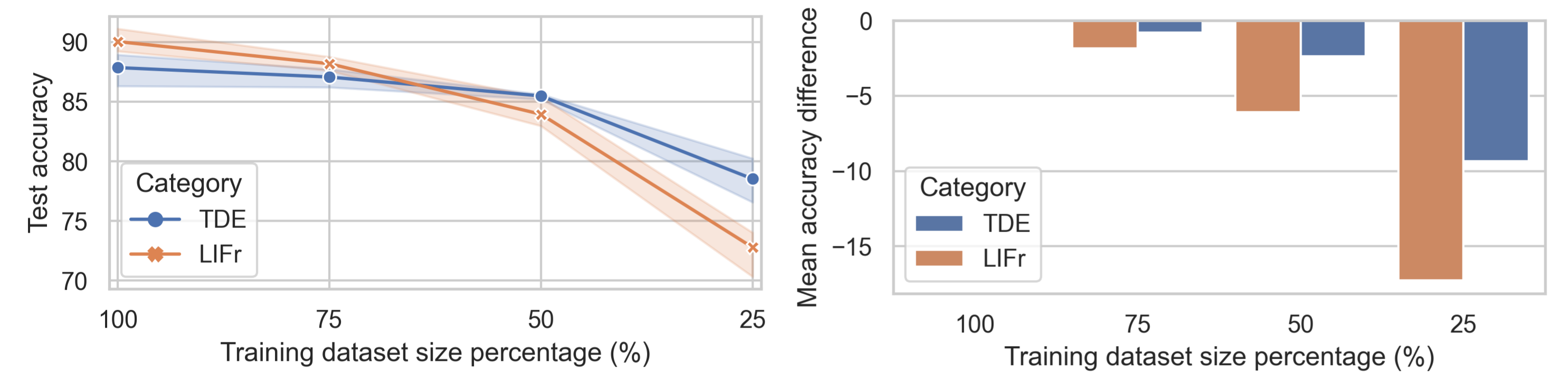}}
\captionof{figure}{Comparison of TDE versus LIF with recurrency in terms of training efficiency. All figures are plotted against a reduction in the training dataset in size percentage. 100$\%$ represents the full training dataset available. The left figure compares the accuracy decrease for both models with the highest accuracy with the decrease in the training dataset. The right figure shows the difference in accuracy against the reduction. TDE architecture is more robust to a reduction in the training dataset.}
\label{fig:results_traineff}
\end{minipage}

\subsection{Interpretation of network parameters}
Given the pre-synaptic compartmentalisation of the TDE cells, every cell has distinct connectivity from the input space. Every cell represents a pair of frequencies from the input. We further correlated these features from the input to every keyword class based on the trained parameters of the network. Specifically, looking at the matrix of connectivity between L1 and L2, represented as $W_2$. We can trace back every cell of the output layer (L2), which corresponds to a different class, to specific features of the input space in the TDE cell. Since every cell is defined by its frequency pair and time constant $\tau_{g}$, we link each keyword class of the dataset with the most relevant frequency pairs of their input at a trained time constant that characterise the time dynamics of the frequency pair.

The first top-left subfigure from Figure \ref{fig:results_footprint} all the possible combinations of the \gls{TDE} cells given its associated frequency pair from the input. In this map, we only represent the highest weight values per keyword class, which contributes the most to the increase or decrease of the neural activity in L2. In the top-right subfigure is represented per keyword class the corresponding distribution of time constants of the \gls{TDE} cells $\tau_{g}$ that mostly contributes to the classification, with the same weight filtering that the previous subfigure. The time constant defines the timescale that better represents the association of the two frequency channels of their input. For instance, a longer time constant here defines, on average, a longer characteristic peak of energy between 2 frequencies, as seen in the word "oh" against a sorter timescale, as we can observe in the word "seven" in the right figure results. 
In the top-left subfigure, we observe that all values are concentrated in the first half. The frequency ranges from 0 Hz to 1500 Hz from a total of 3000 Hz; this frequency band corresponds to the main frequency components found in spoken words. Interestingly, we found small overlap between classes and the \gls{TDE} cells in L1 since each of the points in the frequency bands is represented mainly with a single class/colour in the map. This fact shows that the trained procedure has found distinct features for each class, creating a local population code based on the proposed feature map. 

The bottom-left figure and bottom-right associated table in Figure \ref{fig:results_footprint} compare the trained results with the initial input data analysis of the system. We compared the pairs of frequencies in the top \gls{TDE} cells represented in the top-left figure from the training procedure with the most highly cross-correlated pairs of frequencies per class obtained in the input data pruning procedure of Section \ref{subsec:crosscorr}. We sorted them based on the band proximity between pairs, and we calculated the distance between the individual frequency channels within the network results and between the network results and the input data analysis. The results show a small number of coincidences between these values in the case of the network results against the input data analysis. Therefore, while the input analysis discriminates between the most highly correlated pairs of frequencies, these highly correlated values are not all used in the classification procedure. This is more clearly shown in the bottom-left figure that compares the average distance between frequency pairs calculated in different trained networks and the network results against the input data analysis. The average distance between frequency pairs is significantly higher in the comparative trace of trained features of the network against the input data analysis. Furthermore, this is the case for all the keyword classes, as seen in the table of the figure. 

\begin{minipage}[b]{1\textwidth}
\centering
\includegraphics[keepaspectratio=true,scale=0.5]{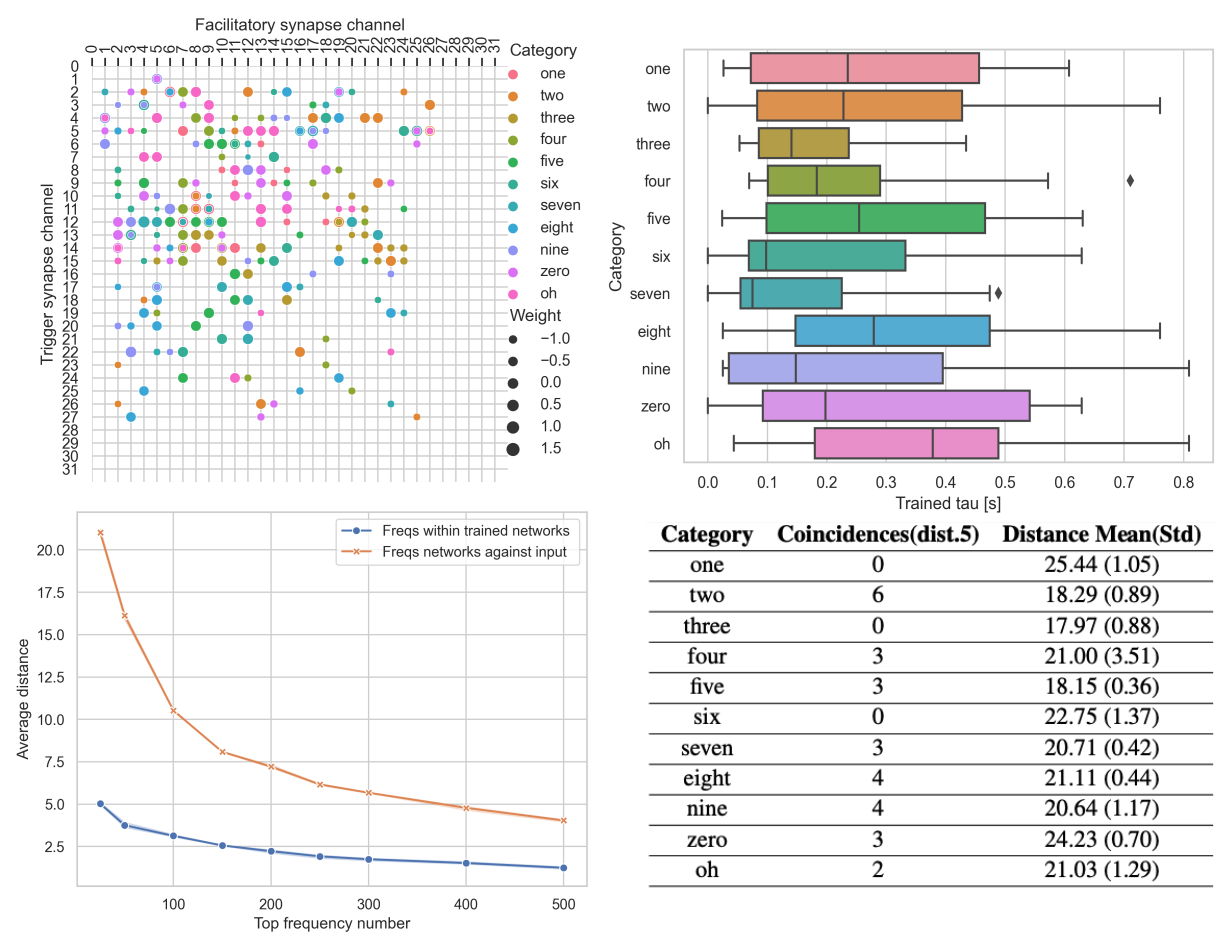}
\captionof{figure}{Frequency pairs and time constants of the TDE network are represented in the context of the classification task. Results were calculated after training output weights $W_2$ and time constants ($\tau_g$) in the \gls{TDE} architecture with all cells in L1. Representation of the top 25 biggest weights in absolute value after training the network. In the top-left figure, each \gls{TDE} cell in the network is represented by a dot in the grid describing its connectivity to the input frequency spectrum. Keyword classes are represented by colour with possibility to overlap between cells. The little overlap found across classes shows a local code of input frequencies for the task. The top-right figure shows the distribution of trained $\tau_g$ per class of the top 25 highest weights in the output layer seen in the left figure. We characterised the temporal dynamics of each keyword in the frequency pair with the \gls{TDE} cell time constant parameter. The bottom-left figure is an analysis of the frequency pair distance between network solutions (blue) and network solutions against initial cross-correlation frequency pairs (yellow). Distance is calculated as euclidean distance with coordinates in the frequency spectrum assigned in the encoding phase. Bottom-right table represents the average distance and coincident frequency pairs between cross-correlation inputs and \gls{TDE} cells with the top 25 values from the other subfigures.}
\label{fig:results_footprint}	    
\end{minipage}

\section{Conclusions}


Considering the speech recognition task, the obtained results motivate using formant decomposition processes found in the literature~\citep{Prica2010, Stanek2013, Laszko2016}. Additional consideration should be taken into account about how to perform this type of computation at the device's edge. Some neuromorphic devices already perform frequency decomposition similar to the biological cochlea that transforms the speech utterances into their frequency content~\citep {liu2010event}. Further work will analyse how to build a posterior processing step to extract the formants, for instance, with resonating filters, as suggested in the literature~\citep{chang1956formant, durrieu2011sparse}.
In this work, we trained \glspl{SNN} using supervised learning methods to maximize classification performance with three different architectures. We found that the proposed dataset contains a significant amount of information in the temporal pattern of spikes that encode the energy of the main frequencies of the speech's formant decomposition. The results show high performance when training TDE and recurrent LIF networks with 89\% and 91\% accuracy, respectively, highlighting the importance of the temporal information in this task. In our experiments, the use of the feedforward LIF network (i.e. without recurrency) is not performing as well in comparison with the other networks. Indeed, recurrency in LIF networks has been shown to improve performance when datasets contain significant information in the time domain~\citep{Perez-Nieves_etal21,Bouanane_etal22, Muller-Cleve2022}.   
While the aim of this work focuses on the comparison between neural models for \gls{SNN} implementations, the hyperparameters have been obtained by exploring a range of values guided by previous insights in~\citep{Muller-Cleve2022}. Future work will explore the most optimized hyperparameters for each network architecture with procedural tuning. Similarly, after analyzing the amount of information in the time pattern versus the rate in this work, we will explore binning and simulation times to exploit the maximal information in the time pattern. Our findings suggest decreasing the time precision in the binning of the dataset to maximize the amount of information related to the spike timing in the pattern.



In the literature, the sparsity of the network is measured as a proxy for energy consumption in neuromorphic processors~\citep{Blouw2020, Sorbaro2020, caccavella_etal23}. The measurements quantify the energy expenditure as the number of computations associated with integrating input spikes into the neuron; this measurement is defined as \gls{SynOps}. Here, we propose to improve the accuracy of the measurement by adding the operations linked to the generation of spikes at every cell. This addition accounts for the energy expenditure to elicit a spike and reset the signals, making it, hypothetically, more accurate in measuring the energy consumption of the hardware implementation. This type of measurement is predictive of the energy consumption~\citep{caccavella_etal23} and relevant to quantify further the advantages of neuromorphic computing with \glspl{SNN}.

In our experiments, the networks differ in the neuron's model used in the hidden layer L1. However, the results regarding the number of synaptic operations are significantly higher even in the consecutive layer L2, where the number and type of neurons are the same for all architectures proposed. This result points to the scalability properties of these architectures since an increase in the depth of the compared LIFrec network would result in further energy consumption due to increased computations deep down in the network hierarchy. These insights are particularly significant if we consider deeper neural network architectures.  
Concerning total computational efficiency, in comparison with the \gls{CuBa-LIF} cell, the \gls{TDE} cell has an extra synaptic compartment and an extra synaptic trace (referred to as the Gain in Figure \ref{fig:methods_tde}). This feature provides richer temporal dynamics, which is useful for temporal pattern recognition. While the complexity of the cell increases, it has far simpler network connectivity with two inputs. In our study, the TDE network presents about 92\% less synaptic operations than the \gls{CuBa-LIF} recurrent network, obtaining an overall advantage in terms of the network's computation efficiency. 
Neuromorphic hardware implementations exploit this gain by using event-driven asynchronous processing. Synaptic operations are only processed in the presence of an event or spike, thus highly reducing the system's total energy consumption~\citep{caccavella_etal23}.

Nevertheless, our metric is an analytical approximation of the real energy consumption which depends on other operations depending on the particular neuromorphic implementation. In digital neuromorphic processors, the state of every leaky compartment should be updated at each algorithmic time step, which can induce a significant constant energy consumption~\citep{lemaire2022}. An asynchronous architecture should compute these traces only when they are relevant to the computation in the presence of an input spike.
This property is already embedded in mixed-signal analog-digital neuromorphic circuits, which makes them a good candidate for the system implementation. There is no algorithmic time step, and the leakage in each compartment of the cell is emulated with the physical properties of the devices (e.g., capacitors)~\citep{Indiveri_etal09,Chicca_etal14b}, resulting in an even higher energy-efficiency in comparison with digital implementations~\citep{Joubert_etal12}.
The neuromorphic community is pushing the boundary in that direction with the design of mixed-signal chips such as Dynaps~\citep{moradi2017scalable} or Texel~\citep{greatorex2024}. 
In these analog-digital neuromorphic circuits, the cost of the complexity of the \gls{TDE} cell is insignificant in comparison with the efficiency presented in this work, with a ~92\% reduction in the synaptic operations. 
Future works will further validate the energy consumption gains of the TDE network by implementing the networks in neuromorphic hardware platforms. Towards that direction, the \gls{TDE} neuron model has already been implemented in Intel's Loihi neuromorphic processor framework \footnote{A. Renner and L. Khacef, "Neuromorphic implementation of \gls{TDE} in Loihi," https://github.com/intel-nrc-ecosystem/models, 2022.}.

We also presented other advantages of using \gls{TDE} neuron cells in an \gls{SNN} architecture aside from the energy efficiency, i.e. data-driven scaling of the network, training data efficiency, and interpretability of the results. First, we analysed a methodology to prune our network based on the cross-correlation of our dataset input frequencies. The \gls{TDE} neuron model transforms into a burst of spikes, the time difference between its input. In certain conditions, its behaviour presents a maximum spiking rate as a coincidence detector when two trains of spikes are highly correlated. By design, we built a layer of \gls{TDE} cells that cover the range of frequency pairs in the input space. Analysing the cross-correlation of the input pairs allows us to identify higher correlated inputs, which will maximise the output of the \gls{TDE} cell. Starting with the less correlated pairs, we were able to prune almost half of the network cells with a very small penalty in the performance (1.22\%). This data-driven methodology has proved to be informative in adapting the size of the network. Applying this methodology, neuromorphic engineers can resize the network based on available resources and the interplay between the performance and efficiency of the system. 
Second, we explore the training efficiency, and the results show a non-statistically relevant decrease in performance, reducing the training dataset to 75$\%$ of its size when using the \gls{TDE} neuron model. The results in training efficiency show that the TDE network is more robust to changes in the dataset and generalises faster to optimal solutions. 
In the comparison with the recurrent LIF, both networks has been balanced in the number of optimisation parameters based on the number of cells cells in $L1$ as similar complexity of the trainable network. In case of the TDE cell, we chose to optimise the time contant of the synaptic computation $\tau_g$ instead of the gain input weight. This parameter describes the timescale of interaction between spikes from different input. 

The \gls{TDE} model used in \glspl{SNN} also contributes to increasing the interpretability of the results, which is a common issue in the training of general Neural Networks~\citep{Zhang2021}. The use of \gls{TDE} cells and their distinctive connectivity applied to shallow networks allow us to apply direct correspondence between the network's trained parameters and the dataset's key features. Every \gls{TDE} in our network is associated with a pair of frequencies through the input connectivity and a specific time constant $\tau_g$. The results from the trained weights in the output layer (L2) present a distinctive code of \gls{TDE} cells (L1) that contributes to each keyword class discrimination. In the same way, we linked the trained $\tau_g$ for each \gls{TDE} cell to the most representative timescale for the coincidence of peaks in the formant frequency spectrum. Therefore, we characterised the dataset's features in the spatial dimension as a set of frequency pairs for each keyword. Through the output code, we also build a distribution of timescales in the formant input space that are distinctive for each keyword speech class.
These distinctive features for classification cannot be obtained through data analysis as we presented in the comparative of those with the analysis of trained parameters. The features obtained from the training network are, on average, very distant in the space of the frequency input bands to the most highly correlated frequencies of the initial analysis. Although the dataset analysis seems insightful to the pruning method, the most informative frequencies of the classification do not relate with the most cross-correlated input frequencies of the dataset. Cross-correlation of input spike train pairs has proven to be informative for network initialisation and training performance. However, the maximal coincidence between these trains of spikes, guaranteed by the cross-correlation measurement, does not extract the most relevant features used for the classification task.

The promising results found in this work point towards the computational efficiency of the \gls{TDE} neuron model for exploiting spatio-temporal patterns of information that are present in all sensory systems. The future prospects in the neuromorphic field of highly efficient digital chips and mixed-signal analog-digital circuits indicate a new generation of energy-efficient hardware. The combination of network architectures that exploit sensory information with highly efficient hardware has the potential to power a new generation of low-power, low-latency devices for full event-driven systems with computations at the edge.


\section*{Conflict of Interest Statement}
The authors declare that the research was conducted in the absence of any commercial or financial relationships that could be construed as a potential conflict of interest.


\section*{Author Contributions}
AP-Z: Conceptualization, Methodology, Software, Validation, Formal analysis, Investigation, Data curation, Writing - original draft, Writing - review and editing, Visualization; LK: Conceptualization, Methodology, Formal analysis, Data curation, Writing - original draft, Writing - review and editing, Supervision; SP: Conceptualization, Methodology, Formal analysis, Writing - review and editing, Supervision; EC: Conceptualization, Methodology, Formal analysis, Writing - review and editing, Supervision


\section*{Funding}
This work has been supported by the European Union's Horizon 2020 MSCA Programme under Grant Agreement No 813713 NeuTouch, the CogniGron research center and the Ubbo Emmius Funds of the University of Groningen.





\bibliographystyle{Frontiers-Harvard} 
\bibliography{bibliography}

\clearpage


\end{document}

%% file: acro_list.tex




\newacronym{SVM}{SVM}{Support-Vector Machine}
\newacronym{SNN}{SNN}{Spiking Neural Network}
\newacronym{BMI}{BMI}{Brain-Machine Interface}
\newacronym{DVS}{DVS}{Dynamic Vision Sensor}
\newacronym{CNN}{CNN}{Convolutional Neural Network}
\newacronym{TDE}{TDE}{Temporal Difference Encoder}
\newacronym{CuBa-LIF}{CuBa-LIF}{Current-Based Leaky Integrate-and-Fire} 
\newacronym{LIF}{LIF}{Leaky Integrate-and-Fire}
\newacronym{HMM}{HMM}{Hidden Markov Model} 
\newacronym{MFCC}{MFCC}{Mel Frequency Cepstrum Coefficient} 
\newacronym{HPO}{HPO}{Hyperparameters Optimization} 
\newacronym{ISI}{ISI}{Interspike Interval} 
\newacronym{mse}{mse}{Mean squared error}  
\newacronym{HMMs}{HMMs}{Hidden Markov Models}
\newacronym{MFCC}{MFCC}{Mel Frequency Cepstrum Coefficients}
\newacronym{BPTT}{BPTT}{Back-Propagation Through Time}
\newacronym{EPSC}{EPSC}{Excitatory PostSynaptic Current}
\newacronym{sEMD}{sEMD}{spiking Elementary Motion Detector}
\newacronym{LIFrec}{LIFrec}{Leaky Integrat-and-Fire with recurrency }
\newacronym{ANN}{ANN}{Artificial Neural Network}
\newacronym{ADAM}{ADAM}{ADAptive Moment estimation}
\newacronym{SynOps}{SynOps}{Synaptic Operations}

%% file: tab/tab_comparison1.tex
\begin{table}
\caption{Comparative table of cross-correlations and equivalent network architectures. While the number of cells in the TDE network is fixed ( based on the preliminary analysis of the data at different levels, see cross-correlation figure for explanation) the number of cells for LIFrec and LIF is calculated to have the same number of connections. The accuracy values are calculated as the mean and standard deviation of three independent learning iterations randomly initialized. The calculated test accuracy is the resultant of the average of the top 25 best test accuracies after training using Backpropagation Through Time (BPTT) to learn the described trainable parameters per network.}
\label{tab:network-comparative}
\begin{tabular}{llllllll}
\multicolumn{1}{c}{\textbf{\begin{tabular}[c]{@{}c@{}}Cross-Corr\\ \textgreater{}=\end{tabular}}} & \multicolumn{1}{c}{\textbf{\begin{tabular}[c]{@{}c@{}}\# TDE \\ cells\end{tabular}}} & \multicolumn{1}{c}{\textbf{Connections}} & \multicolumn{1}{c}{\textbf{\begin{tabular}[c]{@{}c@{}}TDE\\ Accs (std)\end{tabular}}} & \multicolumn{1}{c}{\textbf{\# LIFrec}} & \multicolumn{1}{c}{\textbf{\begin{tabular}[c]{@{}c@{}}LIFrec\\ Accs (std)\end{tabular}}} & \multicolumn{1}{c}{\textbf{\# LIF}} & \multicolumn{1}{c}{\textbf{\begin{tabular}[c]{@{}c@{}}LIF\\ Accs (std)\end{tabular}}} \\ \hline
\textbf{all} & 992 & 12896 & 89.08 (1.25) & 94 & 91.11 (1.07) & 300 & 71.46 (0.44) \\ \hline

\textbf{0.007} & 540 & 7020 & 87.86 (1.12) & 65 & 90.04 (0.80) & 163 & 71.59 (2.72) \\ \hline
\textbf{0.011} & 186 & 2418 & 83.70 (1.55) & 32 & 83.03 (5.14) & 56 & 68.73 (0.34) \\ \hline
\end{tabular}
\end{table}